# Asynchronous Dynamic Bayesian Networks

**Avi Pfeffer**
avi@eecs.harvard.edu
Division of Engineering and Applied Sciences
Harvard University
Cambridge, MA 02138

**Terry Tai**
tai@fas.harvard.edu
Division of Engineering and Applied Sciences
Harvard University
Cambridge, MA 02138

## Abstract

Systems such as sensor networks and teams of autonomous robots consist of multiple autonomous entities that interact with each other in a distributed, asynchronous manner. These entities need to keep track of the state of the system as it evolves. Asynchronous systems lead to special challenges for monitoring, as nodes must update their beliefs independently of each other and no central coordination is possible. Furthermore, the state of the system continues to change as beliefs are being updated. Previous approaches to developing distributed asynchronous probabilistic reasoning systems have used static models. We present an approach using dynamic models, that take into account the way the system changes state over time. Our approach, which is based on belief propagation, is fully distributed and asynchronous, and allows the world to keep on changing as messages are being sent around. Experimental results show that our approach compares favorably to the factored frontier algorithm.

## 1 Introduction

Systems involving multiple autonomous entities that communicate with each other in a distributed, asynchronous fashion are becoming more and more prevalent. Sensor networks are a prominent example of these kinds of systems; teams of autonomous robots are another. In such a system, the entities need to keep track of the state of the system as it evolves. In order to do so, they need to integrate information from throughout the system.

The problem of keeping track of the state of a system as it changes over time is generally called monitoring or filtering. In probabilistic monitoring, the goal is to maintain a probability distribution over the state of the system at each point in time, based on the evidence received up to that point. Distributed, asynchronous systems lead to special challenges to monitoring. Nodes need to update their beliefs about the state of the system autonomously. The timing of updates cannot be synchronized. Instead, each node performs computations at intermittent points in time, and their is no central coordination of the computations. In addition, individual nodes have limited computational power and communication capability.

A further complication is that the systems exist in real time. We cannot stop the evolution of the state as beliefs are being updated. The state continues to change as the monitoring algorithms work. There is therefore a need for asynchronous, message-based algorithms for monitoring and communicating the state of the system, and these algorithms need to be able to work as the state continually evolves.

Techniques for monitoring the state of dynamic systems have typically been centralized and synchronous. A standard representation for dynamic systems is dynamic Bayesian networks (DBNs) [3]. A number of algorithms have been developed for DBNs. Exact inference algorithms [6] compute a complete joint distribution over the state of the system at each point in time. For all but the simplest systems, this joint distribution is too large to be represented and computed, so approximate algorithms are needed. One family of approximate inference algorithms is particle filtering [5], which approximates the distribution over the state of the system by a set of samples. Another approach is the Boyen-Koller algorithm [1] which decomposes the state distribution into a product of factors. All these algorithms rely on centralized computation. One algorithm that can easily be distributed is the factored frontier algorithm [8], which is based on loopy belief propagation and Boyen-Koller. However, even this algorithm is fully synchronous and requires all nodes to coordinate.

Previous work by Crick and Pfeffer [2] showed that loopy belief propagation can provide the basis for communication in distributed, asynchronous settings. They showed that it has several desirable properties. However, their work was based on using static models of the state of the system that do not take into account the way the system changes over time. In their approach, beliefs were based only on current observations, not on history. In monitoring a dynamic system, it is vital to take into account the history of observations in forming beliefs.

In this work we present a framework for distributed, asynchronous probabilistic monitoring that does take into account system dynamics. We call our framework *asynchronous dynamic Bayesian networks (ADBNs)*. In an ADBN, nodes update intermittently at different points in time, and send messages which are then received by other nodes and used when those nodes subsequently update. In order for observations to influence beliefs in the network, nodes maintain a history of beliefs about their state at previous points in time, which are influenced by the observations. These beliefs then influence the belief at the current point in time. Because the time between updates of a node is variable, a continuous time Bayesian network (CTBN) [10] representation of the system dynamics is used. This raises the question of how to convert the CTBN representation into conditional probability tables that can be used in our inference scheme. We present two approaches to answering this question, based on different assumptions about the way variables change their values over time. We present experimental results showing that ADBNs compare favorably to the synchronous factored frontier algorithm.

## 2 Preliminaries

We assume that the reader is familiar with the definitions of Bayesian networks (BNs) [11] and dynamic Bayesian networks (DBNs) [3]. Figure 1 shows two time slices from an example DBN for a fire monitoring domain. In this example there are two rooms. Each room $R_i$ contains a node indicating whether there is a fire in the room. This node depends on whether there was previously a fire in the room and in the adjoining room. There is a node indicating the temperature in the room. This depends on whether or not there is a fire, as well as the outside temperature. Each room has a sensor reading that depends on the temperature. The sensor may be broken.

### 2.1 Loopy Belief Propagation

Pearl's *belief propagation (BP)* [11] is one of the main methods for performing inference in a BN. BP was originally defined for networks without loops, which are undirected cycles. The inference task performed by BP is this: given evidence **e** which is an assignment of values to a subset of variables $\mathbf{E} \subseteq \mathbf{X}$, compute the posterior probability $P(X \mid \mathbf{E} = \mathbf{e})$ for each variable $X$. BP is a message passing algorithm. Each parent $U$ of $X$ sends $X$ a message $\pi_X(U)$, and each child $Y$ of $X$ sends $X$ a message $\lambda_Y(X)$. Based on these, the node $X$ computes $\pi(X)$ and $\lambda(X)$. From these $X$ computes its belief Bel$(X)$ as

$$\mathrm{Bel}(x) = \alpha \pi(x) \lambda(x)$$

where $\alpha$ is a normalizing constant. When the algorithm finishes running, Bel$(X)$ is the desired posterior probability distribution $P(X \mid \mathbf{E} = \mathbf{e})$.

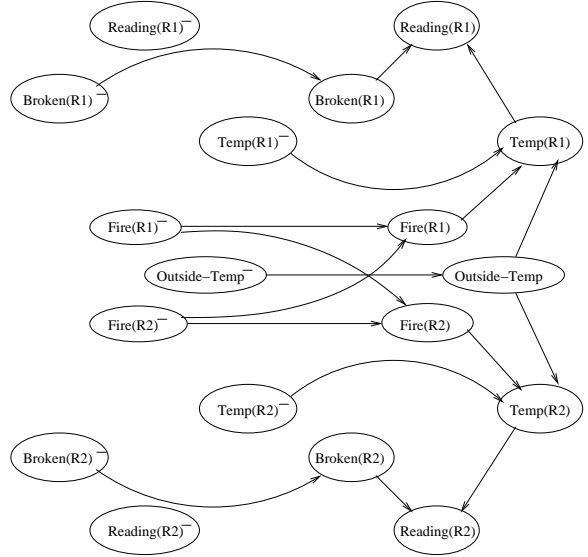

Figure 1: Example DBN

Let the parents of $X$ be $U_1, \ldots, U_j$ and the children of $X$ be $Y_1, \ldots, Y_k$. $\pi(X)$ and $\lambda(X)$ are computed as follows:

$$\pi(x) = \begin{cases} 1 & \text{if } \mathbf{e} : X = x \\ 0 & \text{if } \mathbf{e} : X \neq x \\ \sum_{u_1, \ldots, u_j} P(x \mid u_1, \ldots, u_j) \prod_{i=1}^{j} \pi_X(u_i) & \text{if } X \notin \mathbf{E} \end{cases}$$

$$\lambda(x) = \begin{cases} 1 & \text{if } \mathbf{e} : X = x \\ 0 & \text{if } \mathbf{e} : X \neq x \\ \prod_{i=1}^{k} \lambda_{Y_i}(x) & \text{if } X \notin \mathbf{E} \end{cases}$$

where the notations $\mathbf{e} : X = x$ and $\mathbf{e} : X \neq x$ mean that the evidence **e** assigns value $x$ to $X$ or a different value to $X$, respectively.

Let $V_1, \ldots, V_\ell$ be the children of $U$ other than $X$. The message $\pi_X(U)$ is defined by

$$\pi_X(u) = \pi(u) \prod_{i=1}^{\ell} \lambda_{V_i}(u)$$

Let $W_1, \ldots, W_m$ be the parents of $Y$ other than $X$. The message $\lambda_Y(X)$ is defined by

$$\lambda_Y(x) = \sum_y \lambda(y) \sum_{w_1,\ldots,w_m} P(y \mid x, w_1, \ldots, w_m) \prod_{i=1}^{m} \pi_Y(w_i)$$

Pearl showed that the belief propagation algorithm produces correct posterior probability distributions when run on networks without loops. One may try to run this algorithm on networks with loops, using multiple iterations of passing messages around the network. The resulting algorithm is called *loopy belief propagation (LBP)*. However, in loopy networks the beliefs are not guaranteed to converge, and if they do converge they might not converge to the correct posterior distribution. Nevertheless, empirical results [7, 9] have shown that in a large number of cases, the algorithm converges to approximately correct posterior beliefs in a short amount of time. Thus LBP has emerged as one of the most competitive algorithms for approximate inference in BNs.

## 2.2 The Factored Frontier Algorithm

Let $\mathbf{H}$ be the unobserved variables in a DBN, and $\mathbf{O}$ the observed variables. One of the main inference tasks in DBN is filtering: to compute, at each time point, $P(\mathbf{H}^{(t)} \mid \mathbf{O}^{(1)}, \ldots, \mathbf{O}^{(t)})$. Unfortunately, the cost of performing this task is generally exponential in the number of state variables. One needs to compute a joint distribution over all these variables at every time point. Therefore approximate inference algorithms are needed.

One approximate inference method for DBNs is the *Boyen-Koller algorithm (BK)* [1]. In this approach, the state variables are factored into clusters. Instead of maintaining a joint distribution over all the variables, distributions over the clusters are maintained. The joint distribution is approximated by the product of the cluster distributions. In the original BK algorithm, a join tree is constructed from a two time slice BN, in which there is a cluster containing each of the factors. Inference in this join tree is used to compute the marginal distributions at a time step from the marginals at the previous time step.

Unfortunately, even with the factorization, this join tree computation may be too expensive. An alternative is to use the *factored-frontier algorithm (FF)* [8].

This algorithm performs two approximations. The first approximation, along the lines of BK, is to factorize the state variables into clusters consisting of individual variables, and to approximate the joint distribution over the variables by the product of marginal distributions over the individual variables. The second approximation is to compute the marginals over the individual variables at one time slice from the marginals at the previous time slice using LBP.

## 2.3 Continuous Time Bayesian Networks

*Continuous Time Bayesian Networks (CTBNs)* [10] are a continuous time representation of a dynamic process. CTBNs are based on the theory of Markov jump processes. In a Markov jump process, there is a single state variable that can take on one of $n$ values. The variable begins in a certain state, stays in that state for a random amount of time, then transitions to another state, and so on. The transitions can happen at any point in time. The amount of time the variable stays in a particular state is exponentially distributed.

The dynamics of the process are characterized by an $n$-by-$n$ *intensity matrix* $Q$. The diagonal entries are $-q_i$, and control the rate at which the process leaves state $i$. The amount of time that the process stays in state $i$ is distributed according to the exponential distribution $f_i(t) = q_i \exp(-q_i t)$. The off-diagonal entries of $Q$ describe the transition probabilities between states. When the process leaves state $i$, it enters the next state $j$ with probability $P_{ij} = \frac{q_{ij}}{\sum_{j \neq i} q_{ij}}$. The $q_{ij}$ are scaled so that $q_i = \sum_{j \neq i} q_{ij}$.

If $X$ is governed by a Markov jump process with intensity matrix Q, we use the notation $X^{(t)}$ to indicate the value of $X$ at time $t$. We can derive the conditional probability of $X^{(t)}$ given $X^{(s)}$, where $s < t$, by matrix exponentiation: $P(X^{(t)} \mid X^{(s)}) = \exp(Q(t - s))$.

CTBNs make Markov jump processes applicable to larger and more complex domains, by factoring the state into a number of variables. A CTBN has an associated directed graph. The graph of a CTBN may be cyclic, unlike BNs. Associated with each variable is a set of *conditional intensity matrices (CIMs)*, one for each combination of values of its parents. The CIM for $X$ conditioned on the value $\mathbf{u}$ for the parents $\mathbf{U}$ of $X$ will be denoted by $Q[X \mid \mathbf{u}]$. A CIM specifies an intensity matrix to use for the dynamics of a variable when its parents take on the given value. Given all the CIMs, one can in principle compute a joint intensity matrix where the state space is the cross product of the state spaces of all the variables. This defines the semantics of the CTBN. Of course, this intensity matrix is not computed in practice.

# 3 Asynchronous Dynamic Bayesian Networks

The starting point for our investigation is the factored frontier algorithm. Recall that FF uses loopy belief propagation with the fully factorized belief state. Since LBP is naturally distributed, FF would seem to be an ideal candidate for turning into an algorithm for distributed environments. However, FF is still synchronous. It requires that at every time step all nodes send messages to each other. Furthermore, it requires that at every time step the process stops while the nodes send enough messages for LBP to converge. This means that FF needs a lot of messages to be sent in a very short amount of time, or else ignores changes in the environment that occur during the belief propagation process.

## 3.1 Message Propagation in a Changing Environment

We first discuss the issue of the environment changing while LBP is running. For the purpose of this section, we will still assume a synchronous algorithm. A first attempt towards addressing this issue is to adopt a version of FF in which a full LBP process is not performed at each time step. Instead, each node only sends one message at each time step. So at each time step, nodes would update their beliefs, and propagate their messages to other nodes, to be used at the next time step. Then at the next time step, each node would take in the messages sent at the previous time step, update their beliefs using the DBN dynamics, and propagate new messages to their neighbors. This follows the suggestion of [2], in which LBP is not run to convergence, but rather messages are sent in real time in a changing environment. The difference in the new proposal is that the nodes would also take into account the dynamics of the environment as they update their beliefs.

In fact this proposal would not work as described. The problem is that nodes would only be sending messages forward in time. Messages are sent at one time step to be used at the next. As a result, backward messages would not be propagated. This means that any evidence received from observations would not be taken into account at any nodes that caused the observations. This is not a problem for static models, where the system dynamics are not taken into account. In a static model, there is no notion of messages travelling forwards and backwards in time. In [2], all messages were propagated from one node to another as if time stood still and they were always current. With a dynamic model, on the other hand, messages are time-stamped and travel in time around the unrolled DBN. Therefore backward messages need to be propagated.

A second attempt to make LBP work in a changing environment solves this problem by having nodes propagate backward as well as forward messages. In order to do this, nodes need to maintain historical versions of themselves to receive the backward messages. We call a node $X$ of the DBN a *supernode*, and an instantiation of that supernode $X^{(t)}$ at a particular point in time a *subnode*. In an unrolled fragment of the DBN, there may be many subnodes of the same supernode. In the simplest configuration, a supernode would maintain beliefs about its current state and about its state at the two previous time points. At each point in time, the current subnode will send $\pi$ messages forward to the next time point, and $\lambda$ messages to its parents at the current and previous time points. The $\lambda$ messages will be picked up by the historical subnodes. Meanwhile, the historical subnodes will themselves send $\pi$ messages forward to future time points.

In this scheme, there are two kinds of messages that get passed around, those between supernodes and those between subnodes. Supernodes are the entities in the distributed system that communicate with each other. In contrast, the belief propagation process involves $\pi$ and $\lambda$ messages sent between subnodes. These $\pi$ and $\lambda$ messages are packaged into communications between the supernodes. A supernode then redirects the messages to the appropriate subnode. To avoid confusion, we will call the $\pi$ and $\lambda$ messages "messages", and the messages between supernodes "communications".

This scheme, however, is not an adequate solution. The problem is that messages that are propagated forwards from historical subnodes will never reach the current node and modify the current belief. For example, if a $\pi$ message is sent from the subnode representing time $t-1$ at the current time step, this will be picked up by a subnode representing time $t$ at the next time step. But by the next time step, $t$ is no longer the current time, and the subnode that picks up the message will represent the previous time step. Thus, even though backward messages are passed in this scheme, evidence from observations is never taken into account in the current beliefs. This problem is illustrated in Figure 2.

A better solution, and the one we actually use, builds on the idea of storing historical versions of nodes. Now, though, each supernode performs full-scale inference between its subnodes. This is acceptable since all inference happens within a single supernode, which is a single node in the distributed system. It is the supernodes that communicate with each other. In addition, the inference performed at a supernode is quite simple. The series of subnodes within a supernode form a chain. In order to perform inference amongst them, a single forward and backward pass is sufficient.

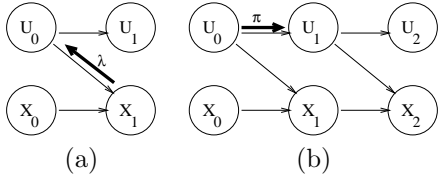

Figure 2: Example illustrating the problem with the second attempt. Thick arrows indicate messages that are sent. (a) Time step 1: a $\lambda$ message is sent from $X_1$ to $U_0$. (b) Time step 2: a $\pi$ message is sent from $U_0$ to $U_1$, but the current time is 2.

To be precise, a supernode performs inference on the fragment of the unrolled DBN consisting of its subnodes. Communications will have been received from other supernodes consisting of current and historical $\pi$ and $\lambda$ messages. These messages from subnodes of other supernodes remain constant throughout the local inference process. The oldest subnode computes a $\pi$ message based on the messages from its non-local neighbors, non-local meaning that they are subnodes of other supernodes (Figure 3 (a)). Then, working forward to the current subnode, each subnode computes a $\pi$ message based on the $\pi$ message from its local parent, the $\pi$ messages from its other parents, and the $\lambda$ messages from its non-local children (Figure 3 (b)). The current subnode then computes a $\lambda$ message from the messages from its non-local neighbors and its local evidence (Figure 3 (c)). Then, working backwards, each node computes a $\lambda$ message from the $\lambda$ message from its local child, the $\lambda$ messages from its other children, and the $\pi$ messages from its non-local parents (Figure 3 (d)). During the backwards pass, each node updates its beliefs based on the $\pi$ and $\lambda$ messages it has received. Also on this pass, each node computes $\pi$ and $\lambda$ messages to send to its non-local children and parents. At the end of the update, the supernode communicates these messages to its neighbors.

Even with this scheme, we still cannot use the beliefs at the current subnode as the beliefs about the system state. Instead we have to use beliefs at a historical subnode. The problem is that the current subnodes will always have just received their first messages at the time they form their first beliefs. Thus they will not have had time to converge to the correct beliefs. In contrast, historical subnodes will have received multiple sets of messages that will have traveled around the loops several times. The belief of a historical subnode about the state of the system at its time point will be much more accurate than the belief of the current node about the current state. We call the subnode used to estimate beliefs about the current state of the system the *reporting subnode*. There is a natural tradeoff here. Older reporting subnodes will have had more chance to converge to the approximately correct beliefs and will be more accurate about the state at their time point. On the other hand, as nodes get older, their beliefs become more stale and become a less accurate reflection of the state at the current time point. In our experiments, we find that the second most current subnode is the best one to use.

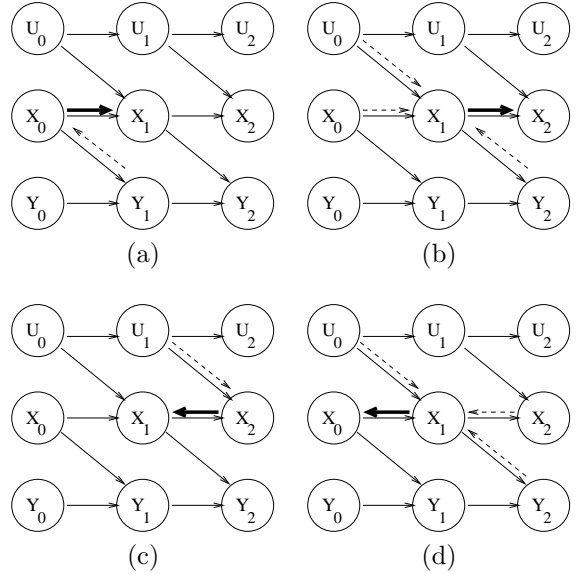

Figure 3: Update of a single supernode. Thick arrows show outgoing messages, and dashed arrows show the messages used to compute them.

One might think that using beliefs about a historical state as a substitute for beliefs about the current state is a drawback of our approach. However, this is unavoidable for any system that updates beliefs in real time. Consider the factored frontier algorithm. Proper beliefs will not be obtained until the loopy belief propagation has had time to converge. By that time, the system will have evolved, so the resulting beliefs will be beliefs about a historical state. It is inevitable that in any system that takes time to process information, the information will be old by the time it has been processed.

### 3.2 Asynchronous Operation

The scheme described so far is still synchronous. All supernodes update at the same time, and subnodes are created at fixed, discrete time intervals. To transform this into an asynchronous scheme, we allow supernodes to update at any point in time. A subnode is created whenever a supernode updates.

In the asynchronous scheme, the amount of time between subnodes of a supernode is variable. In order to represent the dynamics of the system with such variable time lags, we use a continuous time Bayesian network. In a CTBN, each node has a set of conditional intensity matrices, one for each combination of values of the parents. When creating a subnode to participate in the belief propagation process, we need a conditional probability table (CPT) for the subnode. Thus we need to convert the set of CIMs to a CPT for a subnode given its local and non-local parents. We discuss two approaches to this, using different assumptions about the way variables change value over time.

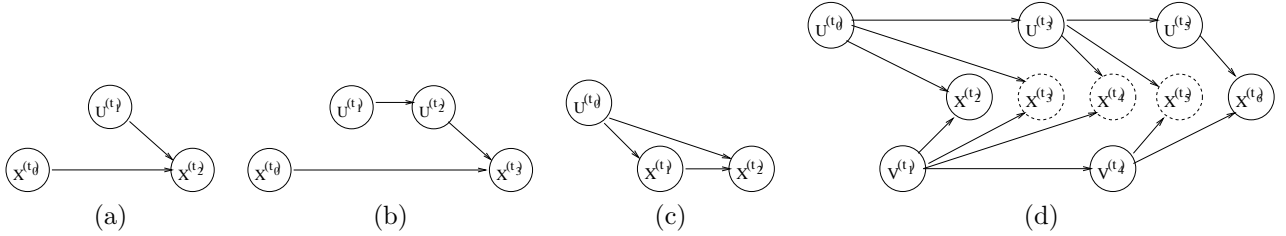

Figure 4: Examples of the two approaches to converting CIMs to CPTs

To make things simple, we will discuss the first approach for a case where a subnode $X^{(t_2)}$ has a single non-local parent $U^{(t_1)}$ and the local parent $X^{(t_0)}$, where $t_0 < t_1 < t_2$. This situation is described in Figure 4 (a). The figure shows the set of subnodes, their timing, and which subnodes are parents of which other subnodes. For each value of $U^{(t_1)}$ and $X^{(t_0)}$, we need a probability distribution over $X^{(t_2)}$. Now we make a strong simplifying assumption, that the value of $U$ was constant throughout the time period $[t_0, t_2]$. This of course is only an approximation, because in reality the value of $U$ might change at any point in the time period. Thus our method has three approximations: this one, and the two inherited from FF, namely, the use of a fully factored representation of the belief state and the use of loopy belief propagation. Because of this assumption, we know that throughout the time period the dynamics of $X$ are governed by the CIM for $X$ conditioned on the value of $U^{(t_1)}$. For a particular value $u^{(t_1)}$ of $U^{(t_1)}$, we can now get $P(X^{(t_2)} \mid X^{(t_0)}, u^{(t_1)}) = \exp(Q[X \mid u^{(t_1)}](t_2 - t_0))$.

Now, consider the case shown in Figure 4 (b), where $U$ has two updates since the last update of $X$, i.e. we have $X^{(t_0)}$, $U^{(t_1)}$, $U^{(t_2)}$ and $X^{(t_3)}$. By the simplifying assumption, the dynamics of $X$ throughout the period $[t_0, t_3]$ are governed by the most recent value of $U$, which is $u^{(t_2)}$. Thus $U^{(t_1)}$ is not the parent of any subnode of $X$. Now consider the opposite case, shown in Figure 4 (c), where $X$ has two updates after an update of $U$, i.e. we have $U^{(t_0)}$, $X^{(t_1)}$ and $X^{(t_2)}$. Since $U^{(t_0)}$ is the most recent update of $X$ before $t_1$ and $t_2$, $U^{(t_0)}$ is a parent of both $X^{(t_1)}$ and $X^{(t_2)}$ and receives $\lambda$ messages from both of them. This makes sense; while $X^{(t_1)}$ provides direct evidence about $U^{(t_0)}$, the $\lambda$ message from $X^{(t_2)}$ will incorporate the influence of $X^{(t_1)}$ and provide additional evidence about $U^{(t_0)}$ which should not be ignored.

All of the above discussion generalizes naturally to cases where children have multiple parents. To be precise, let $U_1, \ldots, U_{n-1}$ be the parents of $X$. Let $t^0$ be the time of the previous update of $X$, and $t^n$ the time of the new update. Let $t^i, i = 1, \ldots, n-1$, be the time of the most recent update of $U_i$ prior to $t^n$. (The $t^i$ are unordered, and we may have $t^i < t^0$. This notation is different from the $t_i$ used before. Whereas before $i$ indicated time ordering, now it indexes parents.) Then the parents of $X^{(t^n)}$ are $X^{(t^0)}$ and $U_i^{(t^i)}, i = 1, \ldots, n-1$. The CPT of $X^{(t^n)}$ is given by

$$P(X^{(t^n)} \mid X^{(t^0)}, u_1^{(t^1)}, \ldots, u_{n-1}^{(t^{n-1})}) = \\ \exp(Q[X \mid u_1^{(t^1)}, \ldots, u_{n-1}^{(t^{n-1})}](t^n - t^0)$$

The second approach to converting the CIMs into CPTs makes a different fundamental assumption. In this approach, we assume that the value of a parent remains constant between updates of the parent. To capture the way a subnode depends on its parents, the period between the last update of a variable and its current update is divided into subperiods, with the beginnings of subperiods corresponding to the times at which parents update. In each subperiod, a particular configuration of values of the parents governs the dynamics of the variable.

Suppose $X$ has parents $U$ and $V$, and let the updates of the variables be $U^{(t_0)}$, $V^{(t_1)}$, $X^{(t_2)}$, $U^{(t_3)}$, $V^{(t_4)}$, $U^{(t_5)}$ and $X^{(t_6)}$. We divide the period $[t_2, t_6]$ between updates of $X$ into the subperiods $[t_2, t_3]$, $[t_3, t_4]$, $[t_4, t_5]$ and $[t_5, t_6]$. In each period, the dynamics of $X$ are governed by the values of the parents at the most recent update prior to the beginning of the period. Thus in $[t_2, t_3]$, the dynamics are governed by $U^{(t_0)}$ and $V^{(t_1)}$; in $[t_3, t_4]$ they are governed by $U^{(t_3)}$ and $V^{(t_1)}$; and so on. Now, if we were to create a CPT directly for $X^{t_6}$, its parents would be all of $U^{(t_0)}$, $U^{(t_3)}$, $U^{(t_5)}$, $V^{(t_1)}$, $V^{(t_4)}$ as well as $X^{(t_2)}$. To avoid a variable having a large number of parents, we decompose the CPT by introducing the subnodes $X^{(t_3)}$, $X^{(t_4)}$ and $X^{(t_5)}$. Each of these subnodes has only one instantiation of each non-local parent as a parent. We then set $P(X^{(t_3)} \mid X^{(t_2)}, u^{(t_0)}, v^{(t_1)})$ to be $\exp(Q[X \mid u^{(t_0)}, v^{(t_1)}](t_3 - t_2))$, and similarly for the other subnodes. The subnodes thus created are not full-fledged participants in the message passing scheme. They do not send $\pi$ and $\lambda$ messages to non-local neighbors. They are used only in local inference. The situation is described in Figure 4 (d). The dashed circles indicate intermediate nodes that are created.

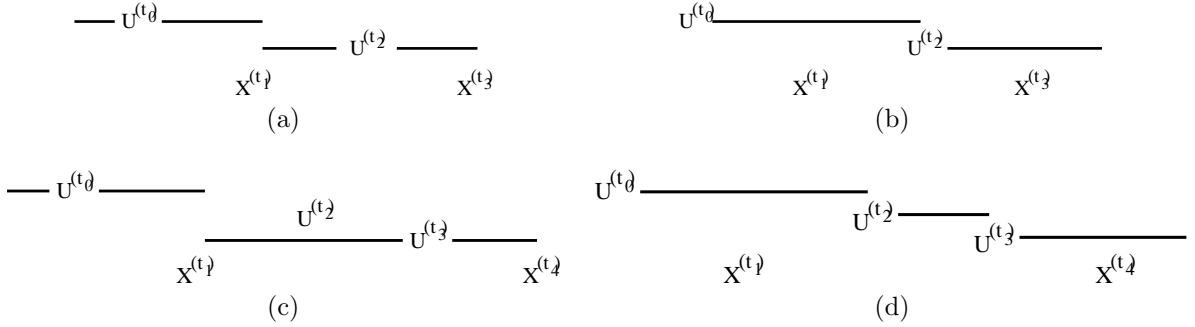

Figure 5: Comparison between the two approaches to converting CIMs to CPTs. The lines indicate during what time each subnode of $U$ governs $X$.

More formally, let $t_1$ be the time of the previous update of $X$, and let $t_n$ be the time of the current update of $X$. Let $t_2, \ldots, t_{n-1}$ be the times of updates of non-local parents of $X$ between $t_1$ and $t_n$. Let the non-local parents of $X$ be $U_1, \ldots, U_m$. For each time $t_i$, $i = 2, \ldots, n$, and each non-local parent $U_j$, let $s_i^j$ be the time of the latest update of $U_j$ prior to $t_i$. Then we create subnodes $X^{(t_i)}, i = 2, \ldots, n$. The parents of $X^{(t_i)}$ are $X^{(t_{i-1})}$ and $U_j^{(s_i^j)}, j = 1, \ldots, m$. The CPT of $X^{(t_i)}$ is given by

$$P(X^{(t_i)} \mid X^{(t_{i-1})}, u_1^{(s_i^1)}, \ldots, u_m^{(s_i^m)}) = \exp(Q[X \mid u_1^{(s_i^1)}, \ldots, u_m^{(s_i^m)}](t_i - t_{i-1}))$$

To compare the two approaches: in many cases one cannot say that one is preferred to the other. For example if we have $U^{(t_0)}$, $X^{(t_1)}$, $U^{(t_2)}$, $X^{(t_3)}$, the two approximations will disagree over which value of $U$ will govern $X$ during $[t_1, t_2]$. According to the first approach (Figure 5 (a)), it will be $U^{(t_2)}$ while for the second approach it will be $U^{(t_0)}$ (Figure 5 (b)). There is no inherent reason to prefer one choice over the other. On the other hand, in some cases the second approach seems more reasonable. If there had been two updates of $U$ before the second update of $X$ (so that we had $U^{(t_3)}$ and $X^{(t_4)}$), then according to the the first approach (Figure 5 (c)) $U^{(t_3)}$ would govern $X$ during $[t_1, t_2]$. This is a poorer choice than $U^{(t_1)}$, which is chosen by the second approach (Figure 5 (d)). On the other hand, the second approach is much more computationally expensive. It does not ensure a constant time per update of a supernode. A constant amount of work per update is important for applications such as sensor networks. Therefore the first approach was used in our experiments.

For an example of how messages propagate in an ADBN, and in particular how evidence is propagated to the current state, consider the situation shown in Figure 6. This example applies to both approaches. In this example $A$ is a parent of $B$, which is a parent of $C$. $C$ is an observed variable. The figure shows the various update times of the different variables.

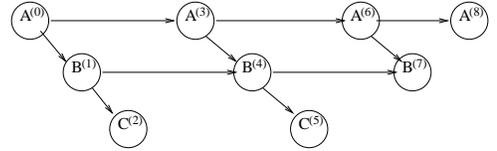

Figure 6: Message propagation example

Notice first that evidence from node $C^{(2)}$ reaches $B^{(4)}$ via $B^{(1)}$. $B^{(4)}$ cannot incorporate evidence from $C^{(5)}$ at the time of its first update, but that is natural since $C^{(5)}$ occurs later. At time 7, the evidence from $C^{(5)}$ reaches $B^{(4)}$ and $B^{(7)}$. Now, at time 6,, evidence from $C^{(5)}$ has not yet reached $B^{(4)}$, so it will not reach $A^{(6)}$. Thus $A^{(6)}$ is unable to incorporate evidence from $C^{(5)}$ at the time of its first update, even though $C^{(5)}$ has already occured. Thus it is incorrect to say that in our scheme, nodes always incorporate all prior evidence into their beliefs. However, as soon as $A^{(8)}$ updates, the evidence from $C^{(5)}$ will reach $A^{(6)}$ and $A^{(8)}$, because it will already have reached $B^{(4)}$ and $B^{(7)}$. Thus eventually, even though it might take some time, evidence is propagated around the network.

### 3.3 Implementation Considerations

If each supernode has to maintain the entire history of its subnodes, the cost of inference within a supern-ode will grow as the process unfolds. To avoid this, we have supernodes phase out their historical subnodes as they become older. There are two possibilities for this. One is for a supernode to maintain a constant number of subnodes. The other is to maintain all subnodes that are younger than a constant amount of time. The advantage of the first method is that it allows a constant inference time at each update, so we use it in our experiments. When a subnode is phased out, its outgoing messages are no longer updated, but remain constant for all time. To allow phased out subnodes to be deleted from memory, all messages are stored by the recipient. That way, if the recipient needs a message from a phased out subnode, it will have the final version of the message.

There is a danger in dropping subnodes too soon. It may prevent evidence from following all paths in the network. In the example in Figure 6, suppose $A_0$ is dropped before $A_6$ updates. Since the evidence from $C_2$ does not reach $B_1$ until $B_4$ updates, it will not reach $A_0$ when $A_3$ updates. Thus at the time $A_6$ updates, since $A_0$ is dropped evidence from $C_2$ will not follow the path $B_1 - A_0 - A_3$. There is a tradeoff here. As the history grows, evidence flows along more paths around the network, but the cost of inference is larger.

So far, we have described a supernode as consisting of all the subnodes for one CTBN variable. In reality, a number of variables might be controlled by the same entity, and no network communication needs to happen between them. Therefore we allow a supernode to consist of all subnodes belonging to a set of variables, and perform local inference, via LBP, amongst all these subnodes every time the supernode updates. In our fire monitoring domain, we have a supernode representing each room.

### 3.4 The Final Algorithm

Our final algorithm, for the case where each supernode corresponds to a single CTBN variable, is shown in Figure 7. This figure shows the process for updating a supernode. This algorithm uses the first approach to converting CIMs to CPTs.

The notation is as follows. $X$ is the supernode being updated. It has parents $U_1, \ldots, U_m$ and children $Y_1, \ldots, Y_\ell$. The subnodes of $X$ are $X_1, \ldots, X_n$. $X_0$ is the subnode of $X$ that has most recently been phased out and is not updated. We assume that a $\pi$ message from $X_0$ to $X_1$ exists, and will never change. We also let $\lambda_{X_{n+1}}(x_n)$ be a vacuous message. The time in which subnode $X_k$ was created is $t_k$.

For each $X_k$, let $U_i^k$ be the most recent subnode of $U_i$ created prior to $X_k$. For each $X_k$, let $Y_i^{k,1}, \ldots, Y_i^{k,h_i^k}$

For $k = 1$ to $n$
$P(X_k | X_{k-1}, u_1^k, \ldots, u_m^k) =$
$\quad \exp(Q[X | u_1^k, \ldots, u_m^k](t_k - t_{k-1}))$
For $k = 1$ to $n$
$$\pi(x_k) = \sum_{x_{k-1}, u_1^k, \ldots, u_m^k} \frac{P(x_k | x_{k-1}, u_1^k, \ldots, u_m^k)}{\pi_{X_k}(x_{k-1}) \prod_{i=1}^m \pi_{X_k}(u_i^k)}$$
$\pi_{X_{k+1}}(x_k) = \pi(x_k) \prod_{i=1}^\ell \prod_{j=1}^{h_i^k} \lambda_{Y_i^{k,j}}(x_k)$
For $k = n$ down to 1
$\lambda(x_k) = \lambda_{X_{k+1}}(x_k) \prod_{i=1}^\ell \prod_{j=1}^{h_i^k} \lambda_{Y_i^{k,j}}(x_k)$
$\lambda_{X_k}(x_{k-1}) =$
$$\sum_{x_k} \lambda(x_k) \sum_{u_1^k, \ldots, u_m^k} P(x_k | x_{k-1}, u_1^k, \ldots, u_m^k)$$
For $i = 1$ to $m$
$\lambda_{X_k}(u_i^k) =$
$$\sum_{x_k} \lambda(x_k) \sum_{x_{k-1}, u_1^k, \ldots, u_{i-1}^k, u_{i+1}^k, \ldots, u_m^k} P(x_k | x_{k-1}, u_1^k, \ldots, u_m^k)$$
For $i = 1$ to $\ell$
$\quad$ For $j = 1$ to $h_i^k$
$\pi_{Y_i^{k,j}}(x_k) =$
$$\pi(x_k) \left( \prod_{\substack{j'=1 \\ j' \neq j}}^{h_i^k} \lambda_{Y_i^{k,j'}}(x_k) \right) \left( \prod_{\substack{i'=1 \\ i' \neq i}}^\ell \prod_{j'=1}^{h_{i'}^k} \lambda_{Y_{i'}^{k,j'}}(x_k) \right)$$
$\text{Bel}(x_k) = \alpha \pi(x_k) \lambda(x_k)$

Figure 7: The final algorithm

be the set of subnodes of $Y_i$ for which $X_k$ is the most recent subnode. The number $h_i^k$ of such subnodes may be zero. We assume that $\pi$ messages from $U_i^k$ to $X_k$, and $\lambda$ messages from $Y_i^{k,j}$ to $X_k$, exist and are held constant throughout the course of the algorithm. For brevity, we have omitted the cases in the computation of $\pi$ and $\lambda$ where the evidence stipulates a value for $X_j$.

The algorithm begins by computing the CPT for each subnode of $X$. In practice this is computed for the subnode once and for all at the time the subnode is created, and does not need to be recomputed on each update. Next comes the forward pass in which $\pi$ values are computed and $\pi$ messages are propagated between local subnodes. Then comes a backward pass. $\lambda$ values are computed for each subnode of $X$, and $\lambda$ messages are passed back to their local parents. In addition, $\pi$ and $\lambda$ messages are computed to send to the non-local children and parents. When computing the $\pi$ message send to the child $Y_i^{k,j}$, the $\lambda$ messages from all other children are used. This is accomplished by incorporating all the messages from other supernodes

$Y_{i'}$ where $i' \neq i$, as well as the messages from other subnodes $Y_i^{k,j'}$ where $j' \neq j$.

## 4 Experimental Results

In our experiments we use the fire monitoring model of Figure 1 extended over a network of 58 rooms, shown in Figure 8. Notice that This room configuration contains some areas with multiple paths between rooms, and some with only a single path. The areas with a single path could be challenging to our algorithm; Crick and Pfeffer [2] showed that redundancy is important to allow LBP to overcome poor beliefs at certain nodes.

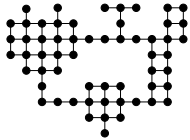

Figure 8: Room structure for experiments

Events were generated using a discrete time generating process, but at a very fine level of granularity, thus approximating continuous time. In the ADBN, nodes updated at much larger time intervals than the time step of the generating process. Nodes updated with a fixed probability at every time step, with typical values for the probability of update being 0.02 to 0.08. The CPTs used in the generating process and for the ADBN updates were derived from CIMs that characterized the behavior of the domain in a natural way. For example, the probability of a fire being generated spontaneously was quite small, but became reasonably large when an adjoining room had a fire, and became larger the more adjoining rooms had fires. More details on the generating model, including CIMs for all the variables, can be found in [12].

We first compared our algorithm with the synchronous factored frontier algorithm. FF updated after a fixed number of time steps of the generating process. To ensure a fair comparison, we made sure that the two algorithms sent the same number of messages. This was achieved by setting the history length for the ADBN equal to the number of LBP iterations for FF, and setting the expected time between updates of each ADBN node equal to the time between updates of FF. In FF, if the number of messages is held fixed, there is a trade-off between the number of LBP iterations and the frequency of update. We found that updating more frequently was more important than running more iterations, so we used two iterations in our experiments.

The results, shown in Figure 4(a), are surprising. In the graph, the $x$-axis represents time; the $y$-axis represents the average negative log likelihood, taken over all rooms, of the true state of the Fire node, at each point in time. The reason for the pattern is that at the beginning of the run, none of the nodes are on fire, and beliefs about this are very accurate. Then one or two fires break out, and since this is quite unlikely the error grows large. Then as time goes on more nodes catch fire, and the network comes to believe that a fire is more and more likely, so the error decreases. The results show that not only did ADBN perform competitively with FF, it outperformed it by a significant margin. It was more quick to believe that a fire had broken out, and as a result its error in the initial stage after a fire had broken out was smaller. The two algorithms converged back to the correct beliefs at about the same time. We found similar results in other experiments comparing the two algorithms. A possible reason for the superior performance of ADBN is that, due to random variation, some nodes will naturally update more frequently than others. These more frequently updating nodes might cause evidence to be propagated more quickly than expected, particularly when there are multiple paths around the network.

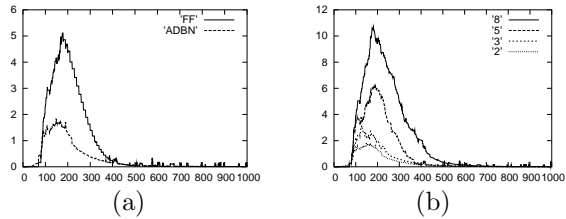

Figure 9: (a) Comparing ADBN to FF; (b) Varying history, using oldest subnode as reporting subnode

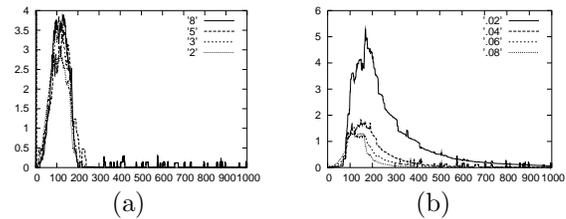

Figure 10: (a) Varying the history, using the second youngest node as the reporting node; (b) Varying the frequency of update

Next, we conducted experiments varying the length of history maintained by the ADBN. Figure 4(b) shows the results for a case where the oldest subnode was used as the reporting subnode, and the update frequency was held constant. The numbers in the legend describe the length of history used. The graph shows that a history of 2 clearly does better. This indicates that it is more important that the reporting node not be stale than that a long history be used. Figure 4(a) shows the results of an experiment in which we used the second youngest node as the reporting node, and varied the length of the history maintained. The graphs are indistinguishable, indicating that there is no advantage to maintaining a longer history. Finally, we experimented with changing the frequency of update of ADBNs. The results are reported in Figure 4(b), where the number in the legend is the probability density of an update occuring at any time point. We see, naturally, that performance improves as the updates become more frequent, though there are diminishing returns.

## 5 Conclusion

We have presented asynchronous dynamic Bayesian networks, a framework for asynchronous, distributed probabilistic reasoning that takes system dynamics into account. Our experimental results show that ADBNs compare favorably with the synchronous factored frontier algorithm. It is important to try to understand better why ADBNs perform so much better than FF. This is a matter for future investigation.

Our framework avoids the need for a centralized controller or for synchronization in monitoring dynamic systems. Our algorithm does not assume that the system stops changing as it updates, but rather performs its message passing in real time as the system evolves. Because the algorithm involves only a fairly small, constant amount of computation at each update, it is suitable for deployment in a system where nodes have limited computational power and communication capability. In a typical run, our algorithm took 0.025 seconds per update on a 3.5 GHz desktop. Furthermore, the frequency of update can be adapted based on the computational resources of a node. This provides a natural way to allocate resources as needed. More powerful nodes can be placed at more vital locations to update more frequently.

This paper has opened up a wide design space for building systems that perform probabilistic reasoning in dynamic and asynchronous environments. We have explored some of the options, but there are many variants left unexplored. For example, we presented two approaches for converting CIMs to CPTs, but others are possible. One could imagine, for instance, letting the value of a parent change linearly between updates. We hope that investigating these design decisions more fully will be a fruitful direction of research.

The next step for ADBNs is deployment in a working network. Sensor networks are being used by a group at Harvard in a medical monitoring domain. This could be an ideal testbed for the ideas in this paper. We would like to extend our framework to monitoring hybrid systems. Currently perhaps the most popular algorithm for this task is Rao-Blackwellised particle filtering [4]. It would be interesting to investigate how to adapt this algorithm to a message-passing framework. Finally, we would like to investigate learning in the message passing paradigm. When a network is deployed in the real world, one rarely has precise knowledge in advance of the workings on the domain. Therefore it would be beneficial to allow the nodes to adapt their model over time, based on the stream of information that they receive.